\documentclass[preprint,12pt]{elsarticle}

\usepackage{amsmath,amssymb,amsfonts}
\usepackage{algorithmic}
\usepackage{graphicx}
\usepackage{textcomp}
\usepackage{xcolor}
\DeclareMathOperator*{\argminB}{argmin}

\usepackage{lineno}

\journal{Information Fusion}

\begin{document}

\begin{frontmatter}



\title{Heterogeneous Federated Learning System for Sparse Healthcare Time-Series Prediction}


\author{Jia-Hao Syu$^a$, Jerry Chun-Wei Lin$^{b,} $\footnote[1]{Corresponding author}\\
$^a$Department of Computer Science and Information Engineering,\\National Taiwan University, Taiwan, f08922011@ntu.edu.tw\\
$^b$Department of Distributed Systems and Informatic Devices,\\Silesian University of Technology, Poland, jerrylin@ieee.org}

\begin{abstract}
In this paper, we propose a heterogeneous federated learning (HFL) system for sparse time series prediction in healthcare, which is a decentralized federated learning algorithm with heterogeneous transfers.
We design dense and sparse feature tensors to deal with the sparsity of data sources. Heterogeneous federated learning is developed to share asynchronous parts of networks and select appropriate models for knowledge transfer.
Experimental results show that the proposed HFL achieves the lowest prediction error among all benchmark systems on eight out of ten prediction tasks, with MSE reduction of 94.8\%, 48.3\%, and 52.1\% compared to the benchmark systems.
These results demonstrate the effectiveness of HFL in transferring knowledge from heterogeneous domains, especially in the smaller target domain.
Ablation studies then demonstrate the effectiveness of the designed mechanisms for heterogeneous domain selection and switching in predicting healthcare time series with privacy, model security, and heterogeneous knowledge transfer.
\end{abstract}


\end{frontmatter}

\section{Introduction}     \label{sec:IN}
With the rapid development of artificial intelligence, deep learning algorithms have been applied in various fields and have shown excellent performance. However, deep learning relies heavily on large amounts of data, and data collection and labeling can be expensive~\cite{DColl1}. In addition, privacy concerns and industry restrictions make data collection and sharing difficult in real-world applications, especially in the healthcare industry~\cite{DColl2}.

To overcome the limited amount of data, transfer learning algorithms are developed to transfer knowledge from domains with rich data or mature models~\cite{TL1}.
Typical transfer learning is applied to domains with the same feature space, which is called homogeneous transfer learning~\cite{Homo}.
In contrast, transferring knowledge from different feature spaces is called heterogeneous transfer, which is still an open problem and a research hotspot~\cite{TTL}.
Moreover, federated learning is advanced transfer learning in which knowledge is iteratively shared, and models are collaboratively trained between clients to obtain general and robust models.
Federated learning is characterized by sharing knowledge rather than data~\cite{FirstFL}.

\begin{figure*}[!htbp] 
  \centering
  \includegraphics[width=1\textwidth]{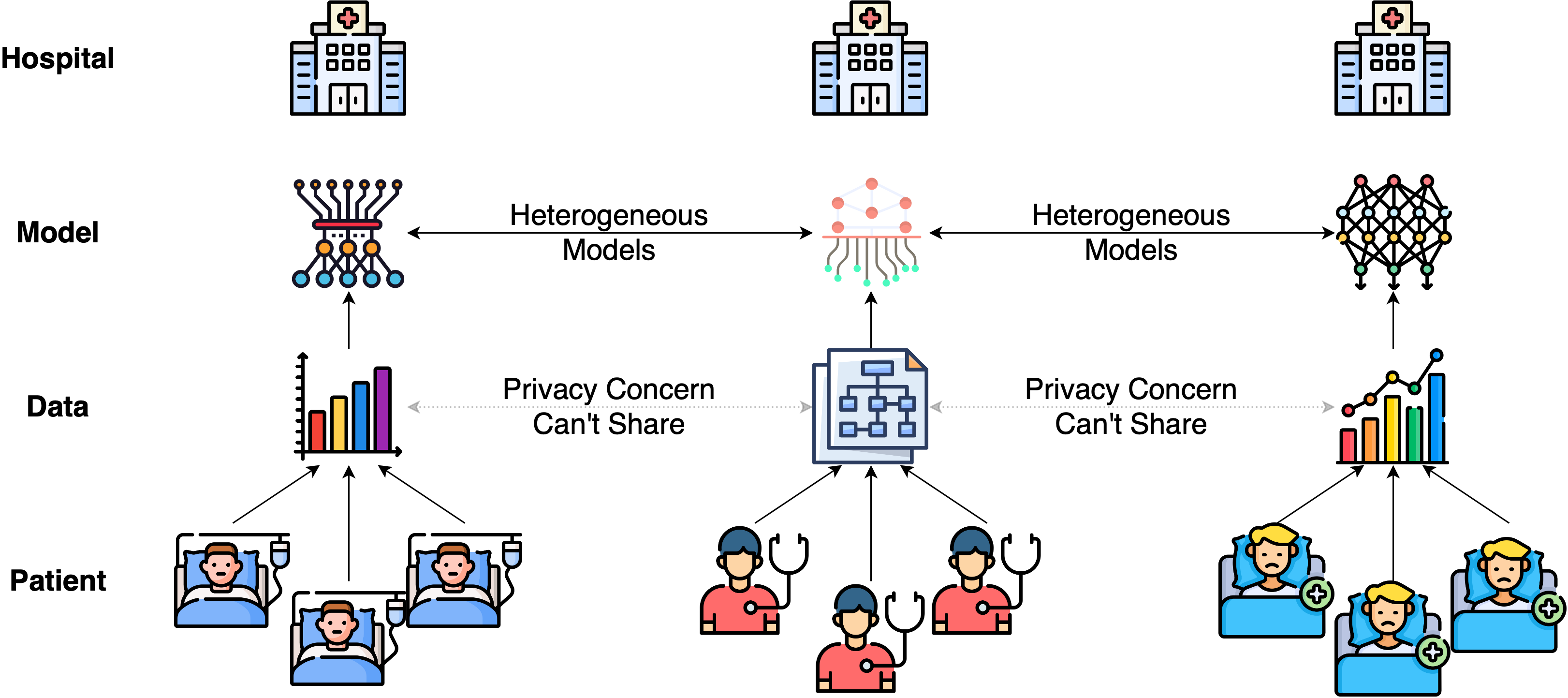}
  \caption{Heterogeneity and security issues in the healthcare industry}
  \label{fig:Flow}
\end{figure*}

The strengths of transfer and federated learning make it suitable for handling privacy-sensitive data, especially in healthcare~\cite{FedHealth, TraHealth}.
Aside from privacy concerns, heterogeneous data are a barrier to implementation in healthcare, which make classical (homogeneous) learning algorithms not suitable and applicable.
As illustrated in Fig.~\ref{fig:Flow}, patient data from different hospitals may be distributed differently, and hospitals may have different equipment and measurements, leading to data mismatches, discrepancies, and sparseness.
These problems lead to heterogeneous data, and privacy issues which further hinder the integration of healthcare data sources.
Therefore, how to securely share knowledge among heterogeneous datasets without sharing data becomes an emerging and compelling question in healthcare research.
However, current research~\cite{HTS1,HTS2,HTS3,BIBE} mainly focuses on predicting healthcare time series using existing machine learning models and places little emphasis on algorithm development, collaborative learning, and the use of multiple datasets while considering privacy and security, which are essential for robust healthcare.

To cope with the sparse and heterogeneous healthcare data, in this paper, we propose a heterogeneous federated learning (HFL) system for predicting sparse time series, validated on healthcare datasets.
HFL is a decentralized federated learning algorithm with heterogeneous transfers.
To deal with the sparsity of data, we develop dense and sparse feature tensors to represent feature-wise and time-series information, respectively.
Several networks and mechanisms are designed in HFL, including global head layers, local embedding layers, prediction layers, heterogeneous domain selection, and switching mechanisms.
The dense feature tensors are the input of global head layers that make preliminary predictions and share knowledge with heterogeneous domains.
The sparse feature tensors are embedded by local embedding layers to distill the temporal information, and the prediction layers then summarize the global and local information to make label predictions.
In addition, a heterogeneous domain selection mechanism has been developed in the proposed HFL to share and select appropriate models for knowledge transfer; furthermore, the switching mechanism is designed to achieve asynchronous sharing with time and computational efficiency.
The contributions and benefits of the proposed HFL can be summarized as follows:

\begin{itemize}
\item Heterogeneous transfer: The target and source domains need not be in the same feature space.
\item Asynchronous: Target and source users can perform asynchronous federated learning to share and select models.
\item Data Privacy: Designing heterogeneous federated learning shares knowledge (model weights) without sharing data.
\item Security: Designing heterogeneous federated learning shares only part of the network, which could prevent data leakage by reverse engineering~\cite{Leak2,Leak1}.
\item Generalization: The proposed system can benefit from other domains of time-series prediction, such as smart factories and smart grids.
\end{itemize}

The experimental results show that the proposed HFL achieves the lowest prediction error and outperforms all benchmark systems on eight out of ten prediction tasks.
The robustness evaluation is also performed for different target and source domains, and the proposed HFL consistently achieves the lowest prediction error on the target dataset with few records, with MSE reduction of 94.8\%, 48.3\%, and 52.1\% compared to three benchmark systems, respectively.
The results demonstrate the effectiveness of the proposed HFL in transferring knowledge from heterogeneous domains, especially in the smaller target domain.
Moreover, ablation studies are conducted to show the effectiveness of the proposed mechanisms in selecting heterogeneous domains and switching, which outperforms random domain selection and always active federated learning.
In summary, the experimental results demonstrate the effectiveness of the proposed HFL system in predicting healthcare time series with privacy, model security, and heterogeneous knowledge transfer.

The following content is organized as follows.
Section~\ref{sec:RW} surveys the literature on transfer learning, federated learning, and healthcare time-series prediction.
Section~\ref{sec:PL} explains preliminary, and Section~\ref{sec:sys} introduce the proposed HFL, including the network design and heterogeneous federated learning mechanism.
Section~\ref{sec:exp} presents and discusses the experimental results, and Section~\ref{sec:con} summarizes the paper.

\section{Literature Review}     \label{sec:RW}
In this section, related research on healthcare time-series prediction is presented in Section~\ref{sec:RW-HP}.
We then introduce transfer learning and federated learning in Sections~\ref{sec:RW-TL} and~\ref{sec:RW-FL}.

\subsection{Healthcare Time-Series Prediction}     \label{sec:RW-HP}
Ever-expanding artificial intelligence technologies are driving the development of smart healthcare, which aims to provide convenient, accurate, and personalized services and includes research focuses such as smart hospitals~\cite{SH}, IoT sensing~\cite{HIoT}, health management~\cite{SHM}, and diagnostic support~\cite{SD}.
However, healthcare services place particular emphasis on privacy and robustness.
How to obtain robust and reliable models with privacy protection is an important issue in health research, for which transfer learning and federated learning have become potential techniques.

Prediction of time-series in healthcare is critical for health management and diagnostic support; therefore, there is a large amount of research.
Kaushik et al. used traditional autoregressive, multilevel perceptron, and long-term memory models to predict patient spending and demonstrated the effectiveness of the ensemble mechanism~\cite{HTS1}.
Kumar and Susan used ARIMA (AutoregRessive Integrated Moving Average)~\cite{HTS2}, and Melin et al. used composite artificial neural networks and fuzzy logic to predict the time-series of COVID-19 cases and deaths~\cite{HTS3}.
Prim et al. developed convolutional network-based feature extractors and self-supervised pre-training systems for predicting SpO2 series (peripheral blood oxygen saturation)~\cite{BIBE}.
Current research mainly uses existing and simple algorithms for healthcare time-series prediction and pays less attention to algorithm design, collaborative learning, and federated learning on multiple datasets with privacy and security considerations.

\subsection{Transfer Learning}     \label{sec:RW-TL}
Machine learning and deep learning rely heavily on big data. However, expensive data labeling and privacy concerns hinder data collection~\cite{TL1}.
With limited data, transfer learning aims to transfer knowledge from data-rich or model-ready domains.
The domain from which knowledge is transferred is called the source domain, while the domain to which knowledge is transferred is called the target domain.
The source and target domains of transfer learning may have different distributions and feature spaces~\cite{TL2, TF3}.
Transferring knowledge from different but related domains can improve the generalization of the model and the efficiency of the training process.

Based on the similarity between the source and target domains, transfer learning can be divided into inductive and transductive transfer learning.
Inductive transfer learning refers to the target and source domains (distribution or feature space) being the same, but the tasks being different~\cite{ITL}.
Multi-task learning is a branch of inductive transfer learning in which two different tasks share the same domain (features or feature extraction) to improve generalization~\cite{MTL1, MTL2}.
Transductive transfer learning refers to the target and source domains being different, but the tasks being the same, with the goal of improving performance through different domain knowledge~\cite{TTL}.

Furthermore, transfer learning can be divided into homogeneous and heterogeneous learning, depending on the feature spaces of the source and target~\cite{TL1}.
In homogeneous transfer learning, the focus is on the target and source domains belonging to the same feature space~\cite{Homo}.
Widely used homogeneous transfer learning involves instance transfer and parameter transfer, i.e., data-level and model-level transfer, respectively.
Heterogeneous transfer learning focuses on target and source domains belonging to different feature spaces. How and what knowledge should be transferred is still an unsolved research problem~\cite{HTL}.

\subsection{Federated Learning}     \label{sec:RW-FL}
Federated learning was first mentioned in 2017~\cite{FirstFL} and focused on collaborative learning without centralized data storage and sharing global models without sharing local data, which is very important for industries with privacy concerns.
Due to the limitation on data sharing, instance transfer is not applicable for federated learning, while parameter transfer is applied for knowledge sharing without data sharing.

Federated learning can be divided into centralized and decentralized~\cite{CDFL}.
Centralized federated learning is controlled by a central server that initiates the aggregation of local knowledge, the updating of the global model, and the dissemination of the latest model~\cite{CFL}.
Decentralized federated learning means that multiple decentralized clients can coordinate knowledge sharing and collaborative model training without a server~\cite{DFL}.

The typical algorithm for federated learning is federated averaging~\cite{FedAvg} (FedAvg).
The client first downloads the global model from the server, trains the model with local data, and uploads the gradient (or model weights) to the server.
The server then averages the gradients (model weights) collected from the clients to update the global model.
Through iterations of local training and global sharing, clients can acquire global knowledge and a robust model without exposing local data.

\begin{figure}[t] 
  \centering
  \includegraphics[width=0.7\textwidth]{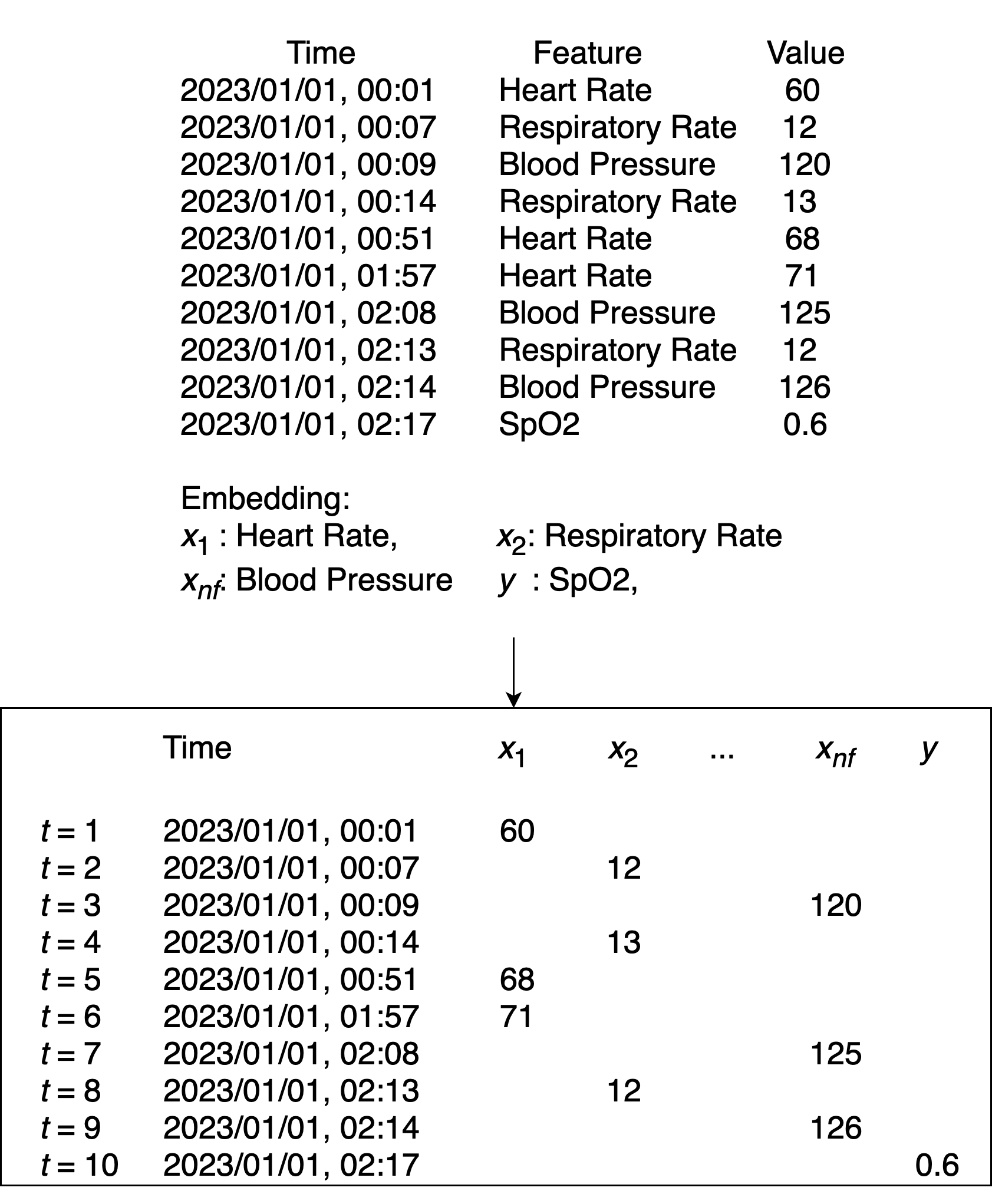}
  \caption{Sparse healthcare data}
  \label{fig:DS}
\end{figure}

\section{Preliminary}     \label{sec:PL}
Due to different measurements and record periods of healthcare information, healthcare datasets are often sparse, i.e., there are many empty or null values in the dataset, as shown in Fig.~\ref{fig:DS}.
In particular, we focus on a dataset that contains only one value (either a feature or a label) at each time $t$.
In addition, the distance between two consecutive time steps may be unequal depending on when the data were collected, complicating data repair algorithms such as interpolation.

Suppose there are $nf$ features and a label.
Let $y_{t}$ be the label at time $t$, and let $x_{i,t}$ be the $i^{\text{th}}$ feature at time $t$, where $i = 1 , \ldots, nf$.
Due to sparsity, only one value is available in [$x_{1,t}, \ldots, x_{nf,t}, y_{t}$] at each time $t$.
To extract sufficient information from the sparse dataset, for each available $y_{t}$, we pack feature tensors for predicting $y_{t}$, including a sparse feature tensor and a dense feature tensor, as detailed in Sections~\ref{sec:PL-SF} and~\ref{sec:PL-DF}.

\begin{figure*}[t] 
  \centering
  \includegraphics[width=1\textwidth]{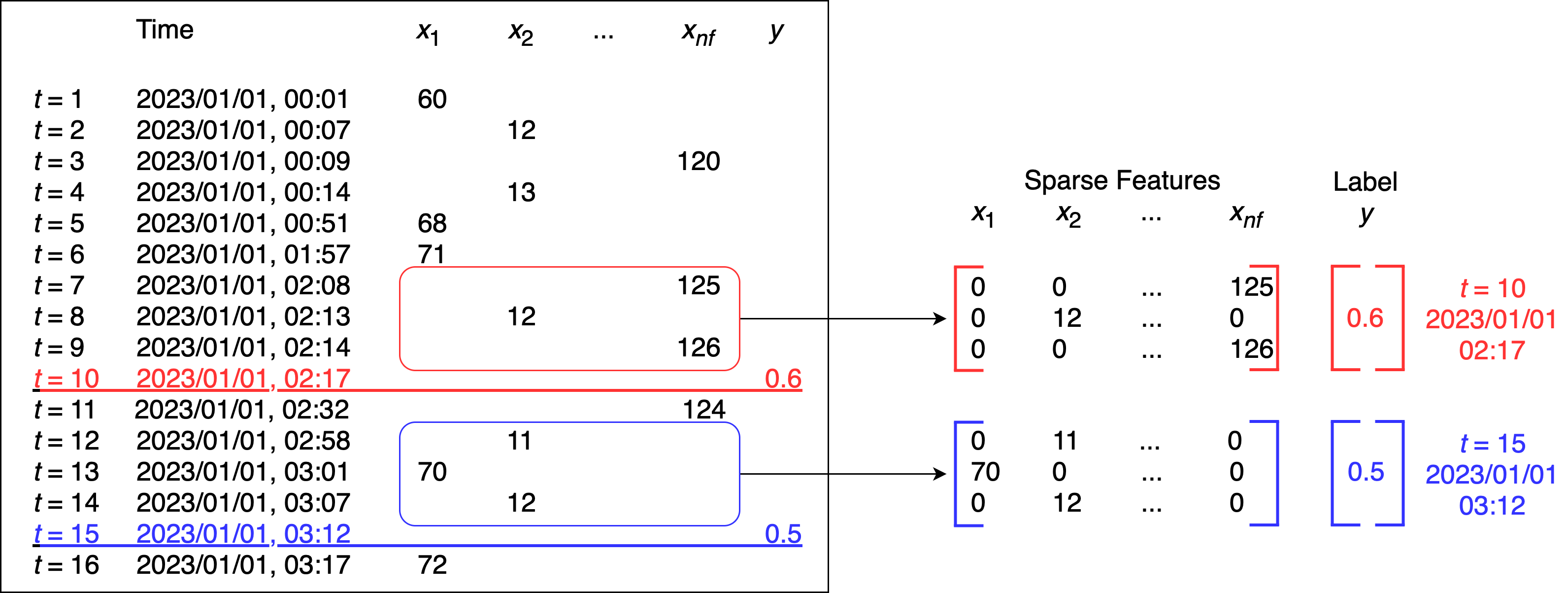}
  \caption{Packing of sparse feature tensors}
  \label{fig:DS-S}
\end{figure*}
\begin{figure*}[t] 
  \centering
  \includegraphics[width=1\textwidth]{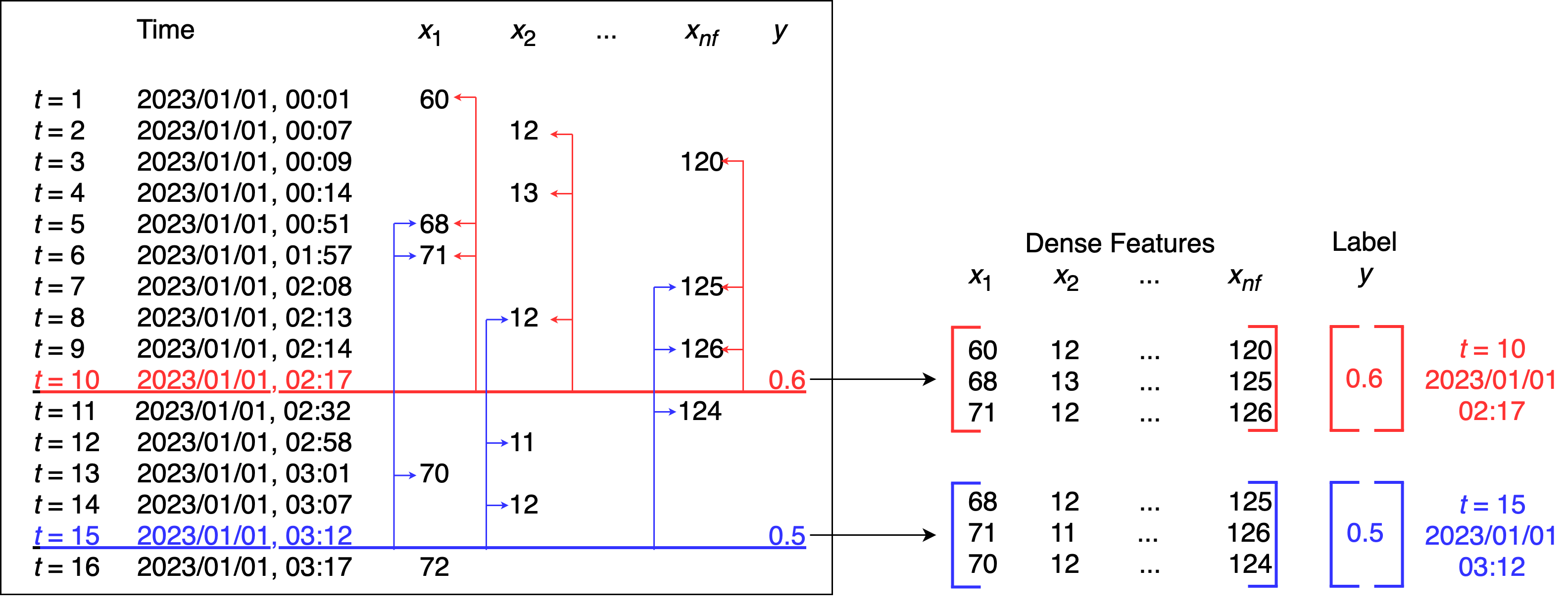}
  \caption{Packing of dense feature tensors}
  \label{fig:DS-D}
\end{figure*}

\subsection{Sparse Feature Tensor}     \label{sec:PL-SF}
To extract the time-series information, we design the sparse feature tensor.
The sketch diagram of the sparse feature tensor is presented in Fig.~\ref{fig:DS-S}.
For an available $y_{t}$, let $X^{S}_{t}$ be the corresponding sparse feature tensor consisting of $nf$ sparse feature vectors, $X^{S}_{1,t}, \ldots, X^{S}_{nf,t}$.
Note that we named a column of feature tensor as a feature vector, and the sparse feature vector $X^{S}_{i,t}$ is defined as:
\begin{equation}
[~ x_{i,t-1}, x_{i,t-2}, \ldots, x_{i,t-w} ~],
\end{equation}
where $w$ is the window size of the feature vector, and $X^{S}_{i,t} \in \Bbb R^{w}$, and $X^{S}_{t} \in \Bbb R^{nf \times w}$.
The sparse features $x$ also result in sparsity for all sparse feature vectors $X^{S}_{i,t}$ and tensors $X^{S}_{t}$.

\subsection{Dense Feature Tensor}     \label{sec:PL-DF}
To extract the feature-wise information, we design the dense feature tensor.
The sketch diagram of the dense feature tensor is presented in Fig.~\ref{fig:DS-D}.
For an available $y_{t}$, let $X^{D}_{t}$ be the corresponding dense feature tensor consisting of $nf$ dense feature vectors, $X^{D}_{1,t}, \ldots, X^{D}_{nf,t}$.
The dense feature vector $X^{D}_{i,t}$ is defined as the last $w$ available values $x_{i,t'}$, where $t'<t$.
The shape of dense feature vectors is $\Bbb R^{w}$, and the shape of the dense feature tensors is $\Bbb R^{nf\times w}$.

For clarity, in the following discussion, we remove the time indices $t$ for which $y_{t}$ is not available, and make the time indices of feature tensors and labels contiguous.
In addition, the variables previously discussed are summarized in Table~\ref{tab:var-p}.

\begin{table}[]
\caption{Variables in Preliminary}
\label{tab:var-p}
\begin{center}
\begin{small}
\begin{tabular}{|l|l|}
\hline
\textbf{Variable} & \textbf{Description} \\
\hline
$nf$ & Number of types of features\\
\hline
$w$  & Window size of feature vectors\\
\hline
$y_{t}$   & Label at time $t$\\
\hline
$x_{i,t}$ & The $i^{\text{th}}$ feature at time $t$, $i=1, \ldots, nf$\\
\hline
$X^{S}_{i,t}$ & The $i^{\text{th}}$ sparse feature vector at time $t$\\
\hline
$X^{D}_{i,t}$ & The $i^{\text{th}}$ dense feature vector at time $t$\\
\hline
$X^{S}_{t}$ & The sparse feature tensor at time $t$\\
\hline
$X^{D}_{t}$ & The dense feature tensor at time $t$\\
\hline
\end{tabular}
\end{small}
\end{center}
\vskip -0.1in
\end{table}
\begin{figure*}[t] 
  \centering
  \includegraphics[width=1\textwidth]{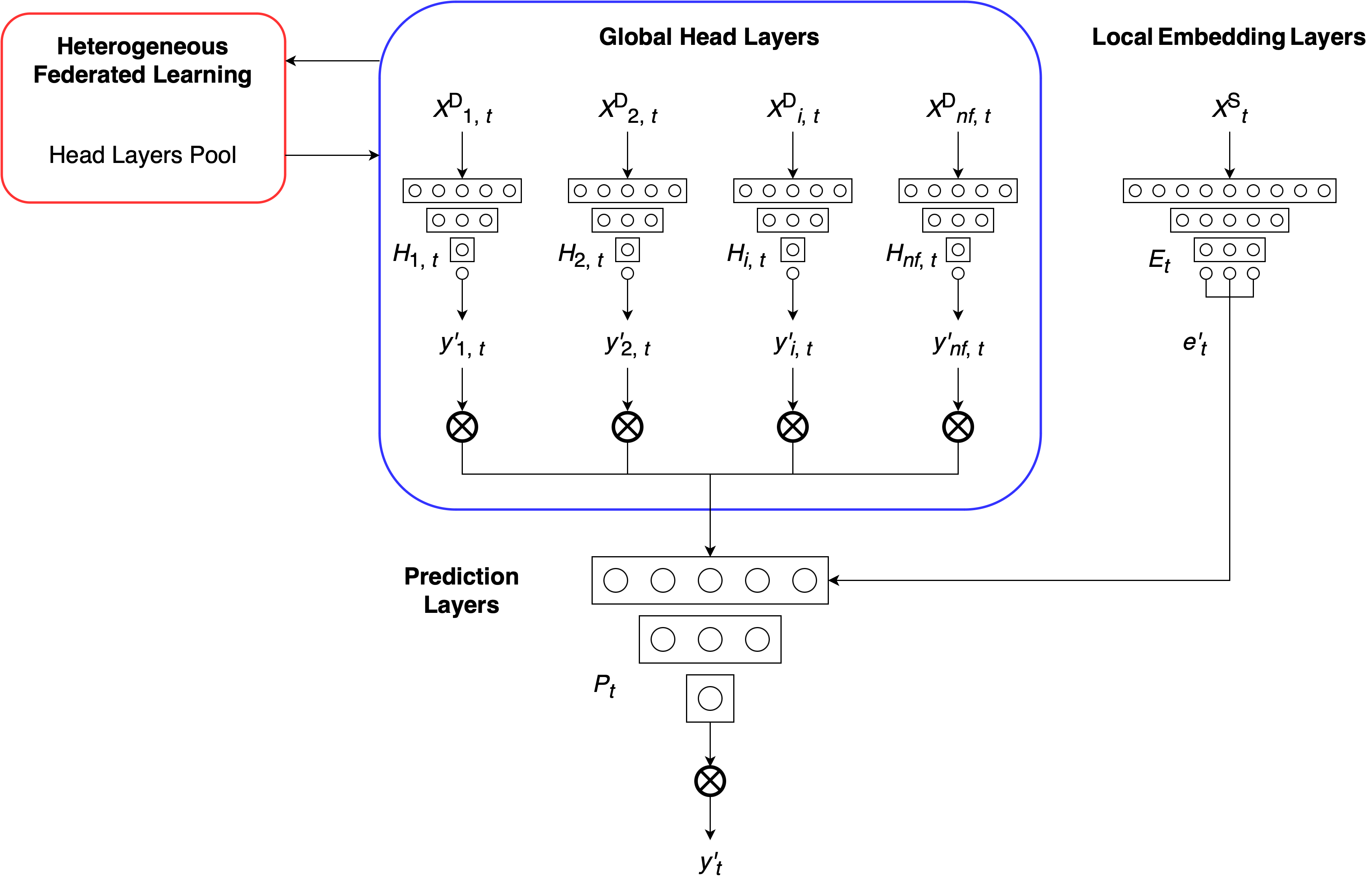}
  \caption{Network design of the proposed HFL}
  \label{fig:ND}
\end{figure*}

\section{Proposed Heterogeneous Federated Learning System (HFL)}     \label{sec:sys}
In this section, we introduce the proposed HFL, including the network design in Section~\ref{sec:sys-ND} and the heterogeneous federated learning mechanism in Section~\ref{sec:sys-HFL}.
In addition, the characteristics of the proposed HFL are discussed in Section~\ref{sec:sys-P}, and the variables of the HFL are summarized in Table~\ref{tab:var-s}.

\subsection{Network Design}     \label{sec:sys-ND}
To independently learn and globally share the feature-wise information, we design the global head layers, which not only process the dense feature tensors, but also enable heterogeneous transferring to obtain global information and knowledge.
On the other hand, to obtain the last information from time-series, we design the local embedding layers to process and embed the sparse feature tensors.
After extracting both feature-wise and time-series information, the prediction layers are designed to summarize the global and local information to make label predictions.
The three major components of the network designed (global head layers, local embedding layers, and prediction layers) are illustrated in Fig.~\ref{fig:ND}.

The global head layers consist of $nf$ deep neural networks $\Bbb H_{1,t}$, $\Bbb H_{2,t}$, $\ldots$, and $\Bbb H_{nf,t}$, where $t$ is the time index.
Each head layers $\Bbb H_{i,t}$ take the corresponding dense feature vector $X^{D}_{i,t}$ as input to make a preliminary prediction of label $y'_{i,t}$, denoted as:
\begin{equation}
\begin{aligned}
y'_{i,t} =~ &\Bbb H_{i,t} ~ X^{D}_{i,t},\\
&i = 1, \ldots, nf.
\end{aligned}
\end{equation}
Each prediction is evaluated by the mean squared error (MSE), expressed as:
\begin{equation}
(y_{t} - y'_{i,t})^{2},
\end{equation}
which is utilized to update the head layers of $\Bbb H_{i,t}$.

The local embedding layers $\Bbb E_{t}$ is a deep neural network that takes the sparse feature tensor $X^{S}_{t}$ as input and embeds it into a $w$-dimensional vector $e_{t}$. 
In other words, local embedding layers $\Bbb E_{t}$ embeds the sparse feature tensor $X^{S}_{t}$ (in a shape of $\Bbb R^{nf \times w}$) into a $w$-dimensional vector to represent the temporal information, denoted as:
\begin{equation}
\begin{aligned}
e_{t} =~ &\Bbb E_{i,t} ~ X^{S}_{t},\\
&X^{S}_{t} \in \Bbb R^{nf \times w}, ~~~ e_{t} \in \Bbb R^{w}.
\end{aligned}
\end{equation}

\begin{figure*}[t] 
  \centering
  \includegraphics[width=0.9\textwidth]{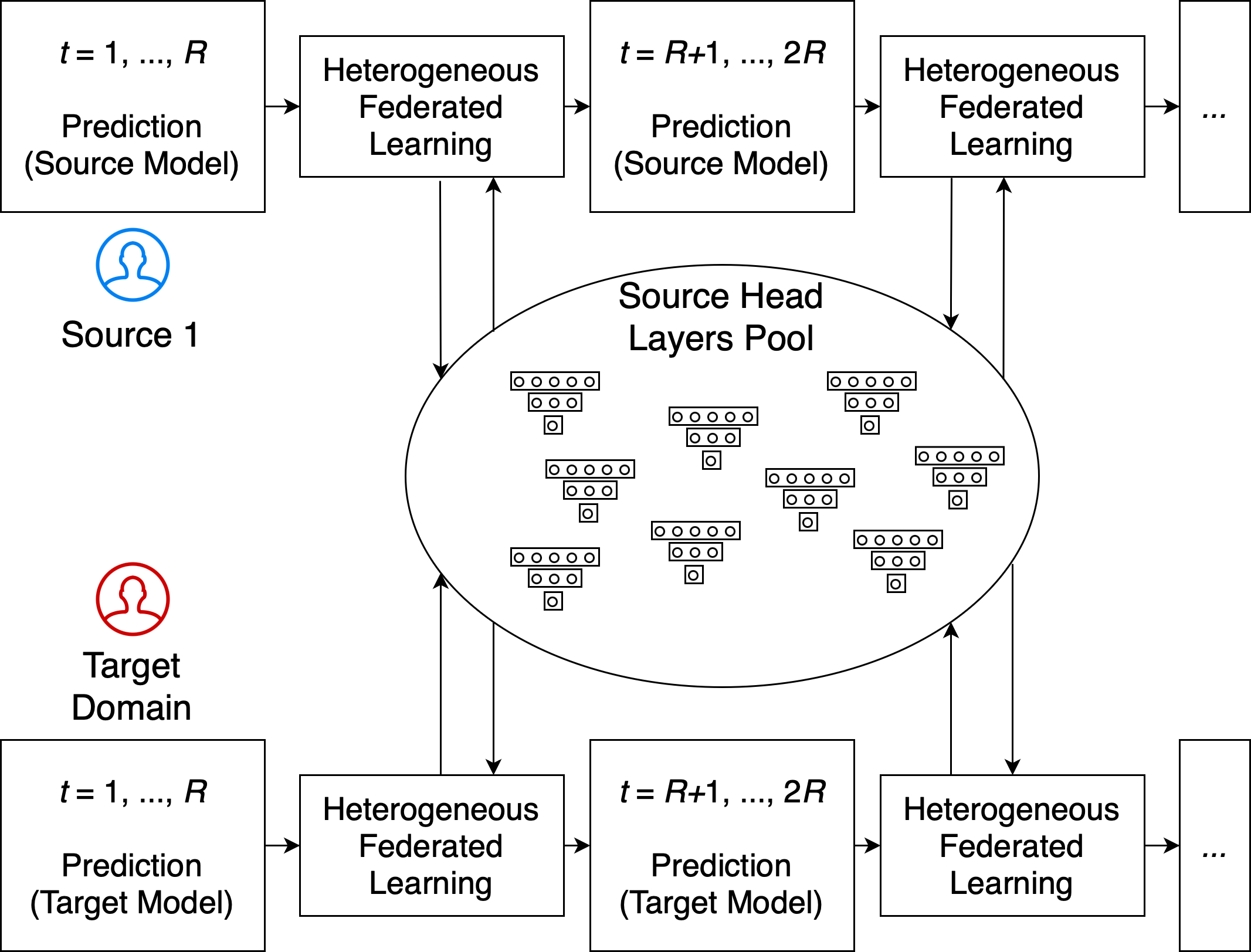}
  \caption{Proposed heterogeneous federated learning mechanism}
  \label{fig:HFL}
\end{figure*}

The prediction layers $\Bbb P$ then aggregate preliminary predictions and the temporal embedded vector to predict the label $y^{'}$, defined as:
\begin{equation}
y^{'}_{t} = \Bbb P_{t} ~ [ y'_{1,t}, \ldots, y'_{nf,t}, e_{t}]^{T}.
\end{equation}
The posterior is the concatenation of $nf$ preliminary predictions and the temporal embedded vector.
The final prediction is also evaluated by MSE, expressed as:
\begin{equation}
(y_{t} - y'_{t})^{2},
\end{equation}
which is utilized to update the prediction layers $\Bbb P_{t}$.
In summary, HFL is designed as a multi-task learning network, consisting of $1 + nf$ tasks evaluated by MSE.

\subsection{Heterogeneous Federated Learning}     \label{sec:sys-HFL}
To overcome the limited amount of data and transfer knowledge from heterogeneous datasets, we design a heterogeneous domain selection to find the most suitable source domain for knowledge transferring.
Additionally, for time and computational efficiency, a switching mechanism is designed to asynchronously execute the heterogeneous federated learning.

In the designed heterogeneous federated learning, there is a pool of source head layers in which users regularly share their weights of head layers $\Bbb H_{1}, \ldots, \Bbb H_{nf}$.
Suppose there are $1 + NS$ users in the environment; one is the target user, and the remaining $NS$ are source users.
The sketch of the two-user environment is shown in Fig.~\ref{fig:HFL}.

The users perform heterogeneous federated learning every $R$ periods (periodically).
During the $R$ periods, the user predicts labels with the same $\Bbb H$, $\Bbb E$, and $\Bbb P$.
After the $R$ periods, the users perform heterogeneous federated learning.
First, for each head $\Bbb H_{i}$, selects the most suitable model from the source head layers pool, called heterogeneous domain selection.
Then, it mixes the weights of $\Bbb H_{i}$, and the selected layers are considered as heterogeneous transfer (referring to FedAvg).
It is worth mentioning that the designed source head layer pool facilitates asynchrony, which means that source users are not necessary to complete computations at the same time (may become a bottleneck due to transmission delay or loss).
Even if some source head layers are not shared into the pool, the proposed HFL can still be executed by the last version stored in the pool.

\begin{table}[t]
\caption{Variables in HFL}
\label{tab:var-s}
\begin{center}
\begin{small}
\begin{tabular}{|l|l|}
\hline
\textbf{Variable} & \textbf{Description} \\
\hline
$\Bbb H_{i,t}$ & Global head layers for the $i^{\text{th}}$ feature at time $t$\\
\hline
$\Bbb E_{t}$   & Local embedding layers at time $t$\\
\hline
$\Bbb P_{t}$    & Prediction layers at time $t$\\
\hline
$y'_{i,t}$ & Preliminary prediction by $\Bbb H_{i,t}$\\
\hline
$e_{t}$ & Temporal embedded vector by $\Bbb E_{i,t}$ and $X^{S}_{t}$\\
\hline
$y^{'}_{t}$ & Final predicted label of $y_{t}$\\
\hline
$R$ & The period of heterogeneous federated learning\\
\hline
$NS$ & The number of source users\\
\hline
$ns$ & The number of source models in the pool\\
\hline
\end{tabular}
\end{small}
\end{center}
\vskip -0.1in
\end{table}

Let the head layers of the target user be $\Bbb H^{T}_{i}$ ($i = 1, \ldots, nf$), and the head layers in the pool be $\Bbb H^{S}_{j}$ ($j = 1, \ldots, ns$, where $ns$ is the number of models in the pool, and equal to the total number of features of $NS$ source users).
The selected model $\widehat{\Bbb H}_{i,t}$ is defined as:
\begin{equation}
\widehat{\Bbb H}_{i,t} = \argminB_{\Bbb H^{S}_{j}, ~j = 1, \ldots, ns} \sum_{l=1,\ldots,R} ( y_{t} - \Bbb H^{S}_{j} ~ X^{D}_{i,t-l+1} ),
\end{equation}
which is the model with the smallest preliminary prediction error in the past $R$ periods.

After selecting the most suitable model, the target head layers are blended as:
\begin{equation}
\Bbb H^{T}_{i} = \alpha~ \widehat{\Bbb H}_{i,t} ~+~ (1-\alpha)~ \Bbb H^{T}_{i},
\end{equation}
where $\alpha$ is a parameter controlling the scale of blending.
Furthermore, we stipulate that the network ($\Bbb P$, $\Bbb E$, and all $\Bbb H_{i}$) is updated by gradient descent every $R$ periods; in other words, each batch of data is in every $R$ time periods.
Therefore, for each $R$ period, the target user updates the network ($\Bbb P$, $\Bbb E$, and all $\Bbb H_{i}$) and then heterogeneously transfers weights of $H_{i}$ through federated learning.

Since heterogeneous federated learning requires additional computation (for model selection), we design a switch for heterogeneous federated learning in the training process.
The switch is designed to perform heterogeneous federated learning only in the epochs where the validation loss has not improved in the last three epochs.
The switching mechanism not only reduces the computational complexity, but also supports the training process when the model is trapped in a local minimum, and provides knowledge and perspectives from the source domain.

\subsection{Properties of HFL}     \label{sec:sys-P}
The advantages of the proposed HFL are summarized as follows.
\begin{itemize}
\item Heterogeneous Transfer: The target and source domains need not be in the same feature space, since the proposed HFL selects heterogeneous models according to the reconstruction errors (representing the similarity of the target and selected features).
\item Asynchronous: Target and source users can perform federated learning asynchronously to share and select models, avoid the cost of synchronization, and increase efficiency.
\item Data Privacy: The designed heterogeneous federated learning shares knowledge (model weights) without sharing data.
\item Security: The designed heterogeneous federated learning only shares part of the network, which could prevent data leakage by reverse-engineering~\cite{Leak2,Leak1}.
\end{itemize}

\section{Experiments and Discussion}     \label{sec:exp}
In this section, we first introduce the data usage and experiment setup in Sections~\ref{sec:exp-DU} and~\ref{sec:exp-ES}.
The prediction evaluation and robustness evaluation are then presented in Sections~\ref{sec:exp-PE} and~\ref{sec:exp-RE}.
Section\ref{sec:exp-AFF} exhibits the ablation studies to demonstrate the effectiveness of the proposed mechanisms.

\subsection{Data Usage}     \label{sec:exp-DU}
In this paper, we evaluate the proposed system using the clinical dataset MIMIC-III ~\cite{MIMIC-2} from PhysioNet~\cite{PhysioBank}.
MIMIC- III is a deidentified health-related dataset collected from thousands of patients in Boston under a data use agreement.
MIMIC- III has two data sources, Carevue and Metavision, which can provide heterogeneous data and characteristics.

In the following experiments, we perform federated learning between the two data sources (which we consider as two hospitals).
Each hospital builds a model to predict its patients' health data.
To improve predictability, both hospitals share their knowledge (model weights) while considering privacy and security (no sharing of patient data).

For each data source, we select the top 5 most numerical features as shown in Table~\ref{tab:Fea}, where M is millions and CF1 to CF5 are selected features from Carevue and MF1 to MF5 are selected features from Metavision.
It can be seen that Metavision is a smaller data source than Carevue.
We also removed patients with insufficient records (less than 1,000).
After this preprocessing of the data\footnote{The origin data can be applied and downloaded from~\cite{MIMIC-2}, and the source code of data preprocessing will be available after acceptance.}, Carevue and Metavision have 4,153 and 2,002 patients, respectively. We divide the patients into 60\% training, 20\% validation, and 20\% testing.
For the prediction label, we choose one feature from the five selected features.
Specifically, we use [CF1, CF2, CF3, CF4] to predict CF5, and [CF1, CF2, CF3, CF5] to predict CF4, etc.

\begin{table}[]
\caption{Selected Features (Top-5 Most Records)}
\label{tab:Fea}
\begin{center}
\begin{small}
\begin{tabular}{|l|l|l|}
\hline
 & \textbf{Feature} & \textbf{Records} \\
\hline
CF1 & Heart Rate & 5.18 M\\
CF2 & SpO2 & 3.42 M\\
CF3 & Respiratory Rate & 3.39 M\\
CF4 & Arterial BP [Systolic] & 2.10 M\\
CF5 & Arterial BP [Diastolic] & 2.09 M\\
\hline
MF1 & Heart Rate & 2.76 M\\
MF2 & Respiratory Rate & 2.74 M\\
MF3 & O2 saturation pulse oximetry & 2.67 M\\
MF4 & Non Invasive Blood Pressure mean & 1.29 M\\
MF5 & Non Invasive Blood Pressure systolic & 1.29 M\\
\hline
\end{tabular}
\end{small}
\end{center}
\end{table}
\begin{table}[]
\caption{Network design of the proposed HFL}
\label{tab:DND}
\vskip 0.15in
\begin{center}
\begin{small}
\begin{tabular}{|c|c|c|c|c|c|}
\hline
\multicolumn{2}{|c}{\textbf{Head $\Bbb H$}} & \multicolumn{2}{|c}{\textbf{Embedding $\Bbb E$}} & \multicolumn{2}{|c|}{\textbf{Prediction $\Bbb P$}}\\
\hline
 \textbf{Layer} & \textbf{Neuron} & \textbf{Layer} & \textbf{Neuron} & \textbf{Layer} & \textbf{Neuron}\\
\hline
Linear & 16 & Linear & 16 & Linear & 32\\
\hline
Sigmoid &  & Sigmoid &  & Sigmoid &\\
\hline
Linear & 256 & Linear & 256 & Linear & 256 \\
\hline
Sigmoid &  & Sigmoid &  & Sigmoid &\\
\hline
Linear & 64 & Linear & 64  & Linear & 16 \\
\hline
LReLU &  & LReLU &  & LReLU &\\
\hline
Linear & 16 & Linear & 16 & Linear & 1 \\
\hline
LReLU &  & LReLU &  & LReLU &\\
\hline
Linear & 1 & Linear & $w$ & Linear & 1 \\
\hline
\end{tabular}
\end{small}
\end{center}
\vskip -0.1in
\end{table}

\begin{table*}[t]
\caption{Prediction Evaluation (MSE) on Metavision Data Source}
\label{tab:PE-M}
\begin{center}
\begin{small}
\begin{tabular}{|c|rr|rr|}
\hline
\textbf{Model} & \multicolumn{2}{c|}{DNN} & \multicolumn{2}{c|}{BIBE} \\
\hline
\textbf{Label} & \multicolumn{1}{c}{Valid} & \multicolumn{1}{c|}{Test} & \multicolumn{1}{c}{Valid} & \multicolumn{1}{c|}{Test} \\
\hline
\textbf{MF1} & 92,533.68 (4) & 61,620.64 (4) & 865.14 (2) & 335.92 (2) \\
\hline
\textbf{MF2} & 1.53 E6 (4) & 2.27 E7 (4) & 48.19 (3) & 42.32 (3) \\
\hline
\textbf{MF3} & 1.75 E8 (4) & 3,505.88 (4) & 1.75 E8 (3) & 777.71 (3) \\
\hline
\textbf{MF4} & 3,621.25 (4) & 2,388.42 (4) & 1,650.23 (3) & 1,428.86 (3) \\
\hline
\textbf{MF5} & 30,022.52 (4) & 10,831.17 (4) & 1,036.10 (2) & 1,084.13 (2) \\
\hline
\multicolumn{5}{c}{}\\
\hline
\textbf{Model} & \multicolumn{2}{c|}{BIBEP} & \multicolumn{2}{c|}{HFL} \\
\hline
\textbf{Label} & \multicolumn{1}{c}{Valid} & \multicolumn{1}{c|}{Test} & \multicolumn{1}{c}{Valid} & \multicolumn{1}{c|}{Test} \\
\hline
\textbf{MF1} & 868.59 (3) & 340.24 (3) & \textbf{852.06} (1) & \textbf{328.30} (1) \\
\hline
\textbf{MF2} & 46.82 (2) & 41.24 (2) & \textbf{46.51} (1) & \textbf{40.87} (1) \\
\hline
\textbf{MF3} & 1.75 E8 (2) & 774.69 (2) & \textbf{1.75 E8} (1) & \textbf{774.11} (1) \\
\hline
\textbf{MF4} & 1,581.88 (2) & 1,357.86 (2) & \textbf{1,368.44} (1) & \textbf{1,147.52} (1) \\
\hline
\textbf{MF5} & 1,119.35 (3) & 1,170.00 (3) & \textbf{561.26} (1) & \textbf{561.97} (1) \\
\hline
\end{tabular}
\end{small}
\end{center}
\vskip -0.1in
\end{table*}

\begin{table*}[t]
\caption{Robustness Evaluation (MSE) on Carevue Data Source}
\label{tab:PE-C}
\begin{center}
\begin{small}
\begin{tabular}{|c|rr|rr|}
\hline
\textbf{Model} & \multicolumn{2}{c|}{DNN} & \multicolumn{2}{c|}{BIBE} \\
\hline
\textbf{Label} & \multicolumn{1}{c}{Valid} & \multicolumn{1}{c|}{Test} & \multicolumn{1}{c}{Valid} & \multicolumn{1}{c|}{Test} \\
\hline
\textbf{CF1} & 1,997.04 (4) & 2,058.45 (4) & 313.84 (2) & 316.34 (2) \\
\hline
\textbf{CF2} & 6,318.39 (4) & 6,485.06 (4) &  18.16 (2) &  19.44 (2) \\
\hline
\textbf{CF3} &   992.63 (4) & 1,006.63 (4) & 120.81 (2) & 122.46 (2) \\
\hline
\textbf{CF4} &   757.51 (4) &   836.83 (4) & 653.84 (2) & 722.90 (2) \\
\hline
\textbf{CF5} &   \textbf{135.91} (1) &   \textbf{149.38} (1) & 196.11 (3) & 209.62 (3) \\
\hline
\multicolumn{5}{c}{}\\
\hline
\textbf{Model} & \multicolumn{2}{c|}{BIBEP} & \multicolumn{2}{c|}{HFL} \\
\hline
\textbf{Label} & \multicolumn{1}{c}{Valid} & \multicolumn{1}{c|}{Test} & \multicolumn{1}{c}{Valid} & \multicolumn{1}{c|}{Test} \\
\hline
\textbf{CF1} & 323.24 (3) & 329.26 (3) & \textbf{306.42} (1) & \textbf{311.50} (1) \\
\hline
\textbf{CF2} &  23.67 (3) &  24.99 (3) &  \textbf{17.89} (1) &  \textbf{18.88} (1) \\
\hline
\textbf{CF3} &  \textbf{96.75} (1) &  \textbf{97.83} (1) & 198.26 (3) & 205.28 (3) \\
\hline
\textbf{CF4} & 654.57 (3) & 722.96 (3) & \textbf{639.26} (1) & \textbf{703.83} (1) \\
\hline
\textbf{CF5} & 199.47 (4) & 213.53 (4) & 160.21 (2) & 172.47 (2) \\
\hline
\end{tabular}
\end{small}
\end{center}
\vskip -0.1in
\end{table*}

\subsection{Experiment Setup}     \label{sec:exp-ES}
In this paper, we take 1 traditional and 2 state-of-the-art systems as benchmarks for comparison.
The traditional benchmark system is a deep neural network (DNN) with four layers (64, 1024, 64, and 1 neuron).
The state-of-the-art systems are BIBE and BIBEP~\cite{BIBE}, which are deep neural networks for SpO2 prediction, and BIBE and BIBEP are systems without and with self-supervised pretraining.
To allow a fair comparison, we change the number of neurons for all systems to have a similar number of parameters (model size). DNN, BIBE, BIBEP, and the proposed HFL have 133,057, 132,234, 132,234, and 131,768 parameters, respectively.

All systems are trained for 50 epochs using the Adam optimizer with a learning rate of 0.01 under the MSE loss.
The save-best mechanism is applied to all systems and saves the model with the lowest error in the validation dataset.
Also, all experiments are repeated 5 times to determine the average performance.
The codes are written in Python and run on Intel(R) Core(TM) i7-9700 CPU @ 3.00GHZ and NVIDIA GeForce RTX 2080 Ti with 11GB GDDR6.
The code of the proposed HFL is available on GitHub\footnote{https://github.com/JiaHao-Syu/Federated-Heterogeneous-Transfer-Learning-HealthCare.git}.

For the proposed system, the detailed network design is shown in Table~\ref{tab:DND}, where LReLU is the activation function of LeakyReLU.
In addition, the variables $w$, $R$, and $\alpha$ are set to 3, 50, and 0.2, respectively.

\subsection{Prediction Evaluation}     \label{sec:exp-PE}
In this section, we perform an evaluation of the prediction of systems predicting MF1, $\ldots$, and MF5 (using the features of [MF2, MF3, MF4, MF5], $\ldots$, [MF1, MF2, MF3, MF4], respectively).
In other words, Metavision is the target domain, and Carevue is the source domain.
The prediction results are shown in Table~\ref{tab:PE-M}, which lists the validation and testing MSEs of the systems.
The values in parentheses are the rankings of the four systems, and the bold values are the best system among the systems (the row).

In Table~\ref{tab:PE-M}, the ranking of validation and testing is the same, which is also not a sign of overfitting.
The experimental results show that the proposed HFL achieves the lowest MSE in all prediction tasks.
HFL achieves the greatest improvement in the prediction of MF5. HFL achieves a test MSE of 561.97, which corresponds to a reduction of 94.8\% (1 - 561 / 10,831.17), 48.3\% (1 - 561 / 1,084.13), and 52.1\% (1 - 561 / 1,170.00) compared to the systems of DNN, BIBE, BIBEP.
Experimental results demonstrate the proposed HFL in healthcare time-series prediction.

\subsection{Robustness Evaluation}     \label{sec:exp-RE}
To evaluate robustness, we run the systems on different target and source domains.
Specifically, we make Carevue the target domain and Metavision the source domain.
Similarly, we run the systems on the prediction of CF1, $\ldots$, and CF5 (using the features of [CF2, CF3, CF4, CF5], $\ldots$, [CF1, CF2, CF3, CF4], respectively).
The prediction results are presented in Table~\ref{tab:PE-C}, listing the validation and testing MSEs of systems.
The values in parentheses are the ranking of four systems, and the values in bold are the best-performing system among the systems (the row).

In Table~\ref{tab:PE-C}, it can be found that the ranking of validation and testing are the same, which shows no signs of overfitting.
The experimental results show that the proposed HFL achieves the lowest MSE when predicting CF1, CF2, and CF4, and takes second and third places when predicting CF5 and CF3, respectively.
The greatest improvement of HFL is predicting CF2, and HFL obtains a testing MSE of 18.88, corresponding to a reduction of 99.7\% (1 - 18.88 / 6,485.06), 2.88\% (1 - 18.88 / 19.44), and 24.4\% (1 - 18.88 / 24.99) compared to the systems of DNN, BIBE, BIBEP.

Comparing the results of Tables~\ref{tab:PE-C} and~\ref{tab:PE-M}, the MSEs in Table~\ref{tab:PE-M} are larger than those of Table~\ref{tab:PE-C}.
We speculate that this is because there are fewer records in Carevue, as shown in Table~\ref{tab:Fea}.
In addition, the data for MF3 is too noisy, leading to extremely high prediction errors on all systems, and simultaneously interferes with HFL predictions on CF3 (the more transfers, the worse the result).
Encouragingly, the proposed HFL performs better on the Metavision data source and always achieves the lowest prediction error.
The phenomenon shows the effectiveness of the proposed HFL in transferring knowledge from heterogeneous domains, especially in the smaller target domain.

\begin{table}[t]
\caption{Ablation study}
\label{tab:AS}
\begin{center}
\begin{small}
\begin{tabular}{|c|rrrr|}
\hline
\text{Model} & \text{No} & \text{Random} & \text{Always} & \text{HFL} \\
\hline
\textbf{CF1} & 311.95 & 311.92 & 312.31 & \textbf{311.50} \\
\textbf{CF2} & 20.67  & 21.23  & 20.10  & \textbf{18.88} \\
\textbf{CF3} & 244.52 & 259.44 & 242.31 & \textbf{205.28} \\
\textbf{CF4} & 704.85 & 704.62 & 704.82 & \textbf{703.83} \\
\textbf{CF5} & 210.47 & 213.26 & 212.52 & \textbf{172.47} \\
\textbf{Aver.} & 298.49 & 302.09 & 298.41 & \textbf{282.39} \\
\hline  
\textbf{MF1} & 328.88 & 329.00 & \textbf{327.99} & 328.30 \\
\textbf{MF2} & 41.71  & 42.21  & 42.06  & \textbf{40.87} \\
\textbf{MF3} & 774.57 & 774.17 & 774.60 & \textbf{774.11} \\
\textbf{MF4} & 1,149.64 & 1,152.26 & 1,149.18 & \textbf{1,147.52} \\
\textbf{MF5} & 565.12 & 566.09 & 567.29 & \textbf{561.97} \\
\textbf{Aver.} & 571.98 & 572.75 & 572.22 & \textbf{570.55} \\
\hline
\end{tabular}
\end{small}
\end{center}
\vskip -0.1in
\end{table}

\begin{figure}[] 
  \centering
  \includegraphics[width=0.7\textwidth]{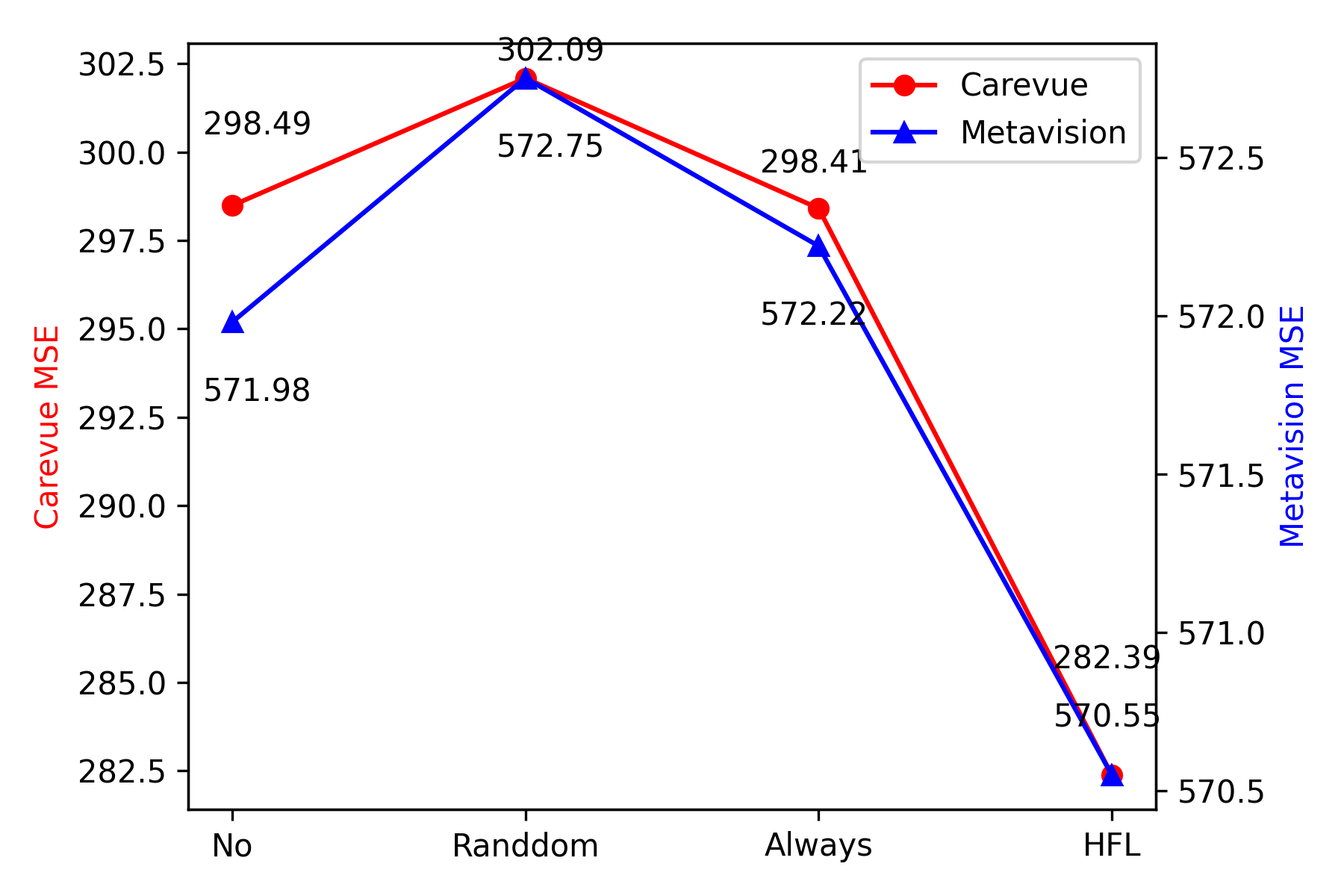}
  \caption{Ablation studies}
  \label{fig:AS}
\end{figure}

\subsection{Ablation Studies}     \label{sec:exp-AFF}
In this section, we perform ablation studies and compare the prediction results of four HFL versions, including HFL-NO, HFL-Random, HFL-Always, and HFL.
HFL- NO uses the same network design as HFL, but without federated learning and heterogeneous transfer.
HFL-Random performs federated learning for each $R$ period (as designed in Section~\ref{sec:sys-HFL}); however, the source headers are randomly selected from the pool.
HFL-Always is the HFL system without the switching mechanism and always performs the federated learning for each $R$ in the training process.

Table~\ref{tab:AS} compares the MSEs of four HFL systems, with the bold values being the smallest among the four versions (the row).
For the Carevue data source, HFL always achieves the lowest MSE.
Similarly, for the Metavision data source, HFL achieves the lowest MSE in four out of five prediction tasks and is slightly higher than HFL-Always in predicting MF1.
The average MSEs of the systems are shown in Fig.~\ref{fig:AS}, where the red and blue lines represent the target ranges of Carevue and Metavision, respectively.
It can be seen that the MSEs of HFL-Random are higher than those of HFL-No, which means that federated learning on heterogeneous domains with random selection affects the prediction.

The MSEs of HFL-Always are lower than those of HFL-Random, demonstrating the effectiveness of the proposed heterogeneous domain selection and transfer.
However, HFL-Always is not necessarily better than HFL-No, and we suspect that the reason is the excessive frequency of transfers.
Comparing HFL-Always and HFL, the results show that HFL significantly reduces MSE and always outperforms HFL-No.
The results also show the effectiveness of the developed switching mechanism in preventing the excessive frequency of transmissions.

\section{Conclusion}     \label{sec:con}
In this paper, we propose a heterogeneous federated learning (HFL) system for sparse time series prediction in healthcare, which is a decentralized federated learning algorithm with heterogeneous transfers.
We develop dense and sparse feature tensors to deal with the sparsity of data sources.
The global head layers share their knowledge with heterogeneous domains and make tentative predictions using dense feature tensors.
The local embedding layers embed the sparse feature tensors to distill the temporal information, and the prediction layers make the final label predictions.
In addition, heterogeneous domain selection and switching mechanisms were developed to asynchronously share and select appropriate models for knowledge transfer.
Experimental results show that the proposed HFL achieves the lowest prediction error and outperforms all benchmark systems on eight out of ten prediction tasks.
Robustness is also evaluated, and HFL consistently achieves the lowest prediction error with MSE reduction of 94.8\%, 48.3\%, and 52.1\% compared to the benchmark and state-of-the-art systems.
The results demonstrate the effectiveness of the proposed HFL in transferring knowledge from heterogeneous domains, especially in the smaller target domain.
Moreover, ablation studies show the effectiveness of the designed mechanisms for selecting heterogeneous domains and switching.
In summary, the proposed HFL is effective in predicting time series in healthcare with privacy, security, and heterogeneous knowledge transfer.
In the future, we will use shared learning to distribute the computational load and enhance the security level.

\bibliographystyle{elsarticle-num}
\bibliography{refs}
\end{document}